\documentclass[letterpaper, 10 pt, conference]{ieeeconf}  

\usepackage{graphicx}
\usepackage{subfigure}

\usepackage{array}
\usepackage{tabularx}
\usepackage{multirow, multicol}
\usepackage[table,xcdraw]{xcolor}
\usepackage{colortbl}

\usepackage{amsmath,amssymb}

\IEEEoverridecommandlockouts                              
\makeatletter
\newcommand*\titleheader[1]{\gdef\@titleheader{#1}}
\AtBeginDocument{\let \st@red@title \@title \def \@title{\bgroup \normalfont \small \flushleft \@titleheader \par \egroup \vskip0.5em \st@red@title}}
\makeatother

\title{\LARGE \bf
Weakly Supervised Recognition of Surgical Gestures
}
\titleheader{Weakly Supervised Recognition of Surgical Gestures. Beatrice van Amsterdam, Hirenkumar Nakawala, Elena De Momi, Danail Stoyanov, In 2019 IEEE International Conference on Robotics and Automation (ICRA).\\
\copyright2019 IEEE. Personal use of this material is permitted. Permission from IEEE must be obtained for all other uses, in any current or future media, including reprinting/republishing this material for advertising or promotional purposes, creating new collective works, for resale or redistribution to servers or lists, or reuse of any copyrighted component of this work in other works.}

\author{Beatrice van Amsterdam$^{1}$, Hirenkumar Nakawala$^{2}$, Elena De Momi$^{3}$, Danail Stoyanov$^{1}$
\thanks{This work was supported by the Wellcome/EPSRC Centre for Interventional and Surgical Sciences (WEISS) (203145Z/16/Z) and the EPSRC (EP/N027078/1, EP/P012841/1, EP/P027938/1, EP/R004080/1).}
\thanks{$^{1}$Beatrice van Amsterdam and Danail Stoyanov are with WEISS and UCL Robotics Institute, Department of Computer Science, UCL, London, UK. {\tt\small beatrice.amsterdam.18@ucl.ac.uk}}
\thanks{$^{2}$Hirenkumar Nakawala is with the Department of Computer Science, Universit\`a di Verona, Verona, Italy.}
\thanks{$^{3}$Elena De Momi is with the Department of Electronics, Information and Bioengineering, Politecnico di Milano, Milan, Italy.}
}

\begin{document}
	
\maketitle

\pagestyle{empty}

\begin{abstract}

Kinematic trajectories recorded from surgical robots contain information about surgical gestures and potentially encode cues about surgeon's skill levels. Automatic segmentation of these trajectories into meaningful action units could help to develop new metrics for surgical skill assessment as well as to simplify surgical automation. State-of-the-art methods for action recognition relied on manual labelling of large datasets, which is time consuming and error prone. Unsupervised methods have been developed to overcome these limitations. However, they often rely on tedious parameter tuning and perform less well than supervised approaches, especially on data with high variability such as surgical trajectories. Hence, the potential of weak supervision could be to improve unsupervised learning while avoiding manual annotation of large datasets. In this paper, we used at a minimum one expert demonstration and its ground truth annotations to generate an appropriate initialization for a GMM-based algorithm for gesture recognition. We showed on real surgical demonstrations that the latter significantly outperforms standard task-agnostic initialization methods. We also demonstrated how to improve the recognition accuracy further by redefining the actions and optimising the inputs. 
\end{abstract}

\begin{keywords}
Classification, Gaussian Mixture Models, robotic surgery, kinematics, surgical gesture recognition
\end{keywords}

\section{Introduction}
Robot-Assisted Minimally Invasive Surgery (RAMIS) is an established practice across a range of surgical specialties, which helps to improve precision of the surgical manipulation and ergonomic comfort of the surgeon \cite{Moorthy2004}. With RAMIS, a large dataset of video and kinematic trajectories of surgical interventions can be recorded from the robotic system, e.g. da Vinci surgical system (dVSS, Intuitive Surgical Inc., CA, USA). Surgical gesture recognition, i.e. segmentation and labelling of surgical action units, by analysing these datasets automatically can be used for multiple purposes, e.g. surgical skills assessment \cite{Reiley2009, Lin2006} and automation \cite{ettehadi2017, fox2017}.

\begin{figure}[t]
	\centering
	\includegraphics[width=0.975\columnwidth]{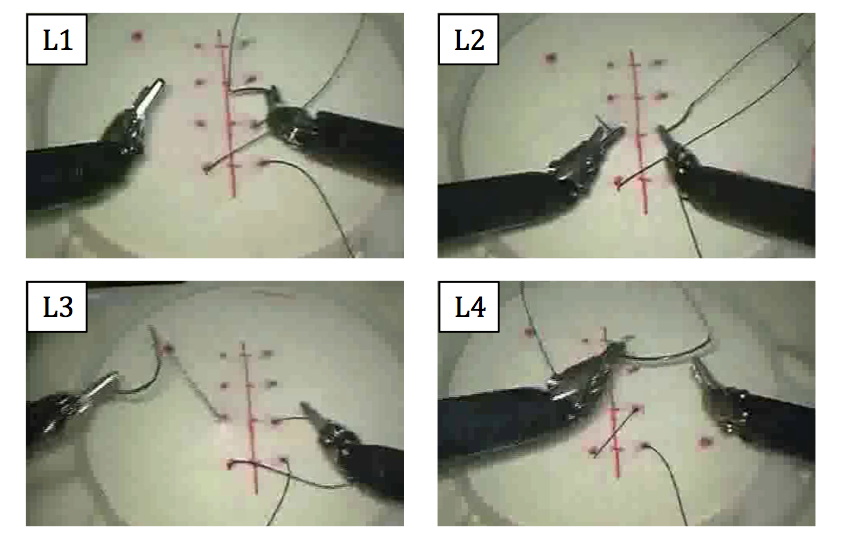}
	\caption[Example of surgemes.]{Example of surgemes \cite{Gao2014}: pushing needle through tissue (L1), transferring needle from right to left (L2), pulling suture with left hand (L3), transferring needle from left to right (L4).}
	\label{fig:surgeme}
\end{figure}

However, automatic gesture recognition is difficult to implement robustly due to the complexity of surgical tasks and the variability in users' actions and patient-specific anatomy \cite{Cao1996}. A number of approaches have been proposed to address this problem. Classical approaches are based on statistical models such as Gaussian Mixture Models (GMM) \cite{Krishnan2018}, Hidden Markov Models (HMM) \cite{Tao2012,Varadarajan2009} and Conditional Random Fields (CRF) \cite{Tao2013, Mavroudi2018}. More recently, deep learning techniques have also been employed, providing the current state-of-the-art results \cite{Twinanda2017,Lea2016a,Lea2016b}. The challenge with most of these methods is that they use manual annotations, which are costly, time consuming and subjective when generated by multiple participants. Due to subjectivity and smooth transitions between gestures, boundaries between consecutive gestures are often not clearly defined.
Unsupervised methods, that automatically learn the segmentation criterion from the data, have been developed to overcome these limitations \cite{Krishnan2018}, \cite{Despinoy2016}, \cite{Fard2017}. However, they typically perform less well than methods 
trained on labelled information, especially on data with high variability such as surgical trajectories. Hence, the potential of weak supervision could be to improve unsupervised learning while avoiding manual annotation of large datasets.

The aim of this paper is to propose a new weakly supervised approach for surgical gesture recognition, that allows to retain the amount of annotations of surgical gestures limited to very few demonstrations. In particular we used at a minimum one expert demonstration and its ground truth annotations to generate an appropriate initialization for a GMM-based unsupervised recognition algorithm, in order to improve upon standard task-agnostic initialization methods, such as random or K-means initialization \cite{Blomer2013}. 
We focused on recognition of \textit{surgeme} units (Fig.\ref{fig:surgeme}), the shortest ``surgical motion unit with explicit semantic sense" \cite{Despinoy2016} (e.g. \textit{grasping the needle}, \textit{pulling the suture}, etc.). We validated our algorithm on the JIGSAWS dataset \cite{Gao2014, Gao2017}, featuring suturing demonstrations collected from eight surgeons with different skill level using the dVSS.

\section{Related work}

\textit{GMM-based methods}: 
this group of works segment robot trajectories 
into action classes by fitting a GMM onto the available samples.
In \cite{Lee2014} the number of mixture components was chosen using the Bayesian Information Criterion (BIC) and the fitting was initialized using the K-means clustering algorithm.
\cite{Krishnan2018} presented a multi-level clustering approach for identification and pruning of segmentation points, which is based on a Dirichlet Process GMM (DPGMM), i.e. a mixture model where the number of clusters is determined by a DP. 
This work was extended integrating features extracted with deep neural networks from the video data \cite{Murali2016}, improving the recognition accuracy. 

\textit{Heuristic initialization}: a number of studies make use of heuristics to create an initial segmentation of the data. 
Examples include heuristics based on Zero Crossing Velocity \cite{fod2002}, jerk profiles \cite{rohrer2006} or trajectory curvature \cite{lioutikov2017}. However, 
the same heuristic could be not appropriate to explain every part of the data, and the output is often over-segmented \cite{lioutikov2017}. 

\textit{Weakly supervised methods}:
only few weakly supervised approaches have been developed for surgical action recognition  \cite{quellec2014}, \cite{padoy2012}, \cite{lalys2011}. Some works, however, assume that the actions follow a pre-defined order, the goal is to find the action boundaries  \cite{padoy2012}, \cite{lalys2011}. In addition, these methods were only applied to recognise surgical phases, which represent high-level surgical states, using only the video data and lack the recognition of low-level surgeme units. 

\section{Methods}

Our approach relies on classical GMM clustering. Multiple reasons make GMM an appealing method for gesture recognition, such as performing simultaneous segmentation and classification, where one task does not rigidly influence the other, as in sequential approaches. GMM is intuitive because action classes are represented through independent means, covariance matrices and weights. Finally, the fuzziness at the segment boundaries is modelled through Gaussian intersections.

Notation: vectors are represented in bold lowercase letters (e.g. \textbf{x}), matrices are represented in bold capital letters (e.g. \textbf{A}) and scalars are represented in italic letters (e.g. \textit{t}).

\subsection{Data pre-processing}

We used JIGSAWS \cite{Gao2014}, a public dataset comprising video and kinematic data captured at 30 Hz from the dVSS during multiple demonstrations of elementary surgical tasks, which were performed on phantoms by eight surgeons with different robotic surgical experience (expert, intermediate, novice). JIGSAWS also contains manual annotations describing the ground truth segmentation of each demonstration into action classes.

We tested our algorithm on the kinematic data recorded from the two Patient Side Manipulators (PSM1 and PSM2) of the dVSS \cite{Kazanzides2014}. The motion of each arm is described by a local frame attached at its end-effector using 19 kinematic variables, including Cartesian positions, a rotation matrix, linear velocities, angular velocities and a gripper angle. 

The pre-processing pipeline of \cite{Despinoy2016} was implemented:
\begin{itemize}
	\item The rotation matrix \textbf{R} describing the end-effector orientation with respect to the robot base is converted into a more compact quaternion representation \textbf{q}, reducing the state vector to 14 variables for each arm.
	\item All the trajectories are smoothed with a low-pass filter with cut-off frequency $f_c=1.5$ Hz in order to minimize the measurement noise.
	\item All the trajectories are normalized to zero mean and unit variance, in order to enable a fair comparison between signals with different unit of measure.
\end{itemize}

\begin{table}[t]
	\vspace{0.3cm}
	\centering
	\footnotesize
	\caption[Kinematic feature vector.]{Kinematic feature vector.}
	\normalsize
	\label{table:kinem_variab}
	\begin{tabular}{|c|c|c|}\hline
		\textbf{Indices} 							&\textbf{Description of variables}\\\hline
		1--3										&Right PSM tool tip position ($xyz$)\\\hline
		4--7										&Right PSM tool tip orientation quaternion ($\textbf{q}$)\\\hline
		8--10										&Right PSM tool tip linear velocity ($\dot{x}\dot{y}\dot{z}$)\\\hline
		11--13										&Right PSM tool tip rotational velocity ($\dot{\alpha}\dot{\beta}\dot{\gamma}$)\\\hline
		14											&Right PSM gripper angle ($\theta$)\\\hline
		15--17										&Left PSM tool tip position ($xyz$)\\\hline
		18--21										&Left PSM tool tip orientation quaternion ($\textbf{q}$)\\\hline
		22--24										&Left PSM tool tip linear velocity ($\dot{x}\dot{y}\dot{z}$)\\\hline
		25--27										&Left PSM tool tip rotational velocity ($\dot{\alpha}\dot{\beta}\dot{\gamma}$)\\\hline
		28											&Left PSM gripper angle ($\theta$)\\\hline
		29--32										&Euclidean distance signals ($d_x,d_y,d_z,d$)\\\hline
	\end{tabular}
\end{table}

Additionally, four signals representing the distance between the two end-effectors along the three orthogonal axes ($d_x,d_y,d_z$) and their absolute Euclidean distance ($d$) are generated, in order to include information about the relationship between the two manipulators, resulting in a state vector $\textbf{x}(t) \in R^p$ of $p=32$ variables. Table \ref{table:kinem_variab} describes in detail the variables included in the kinematic feature vector.

Finally, the trajectories are subsampled from 30 Hz to 10 Hz for faster computation time.

\begin{figure*}[t]
	\vspace{0.5cm}
	\centering
	\includegraphics[width=0.85\textwidth]{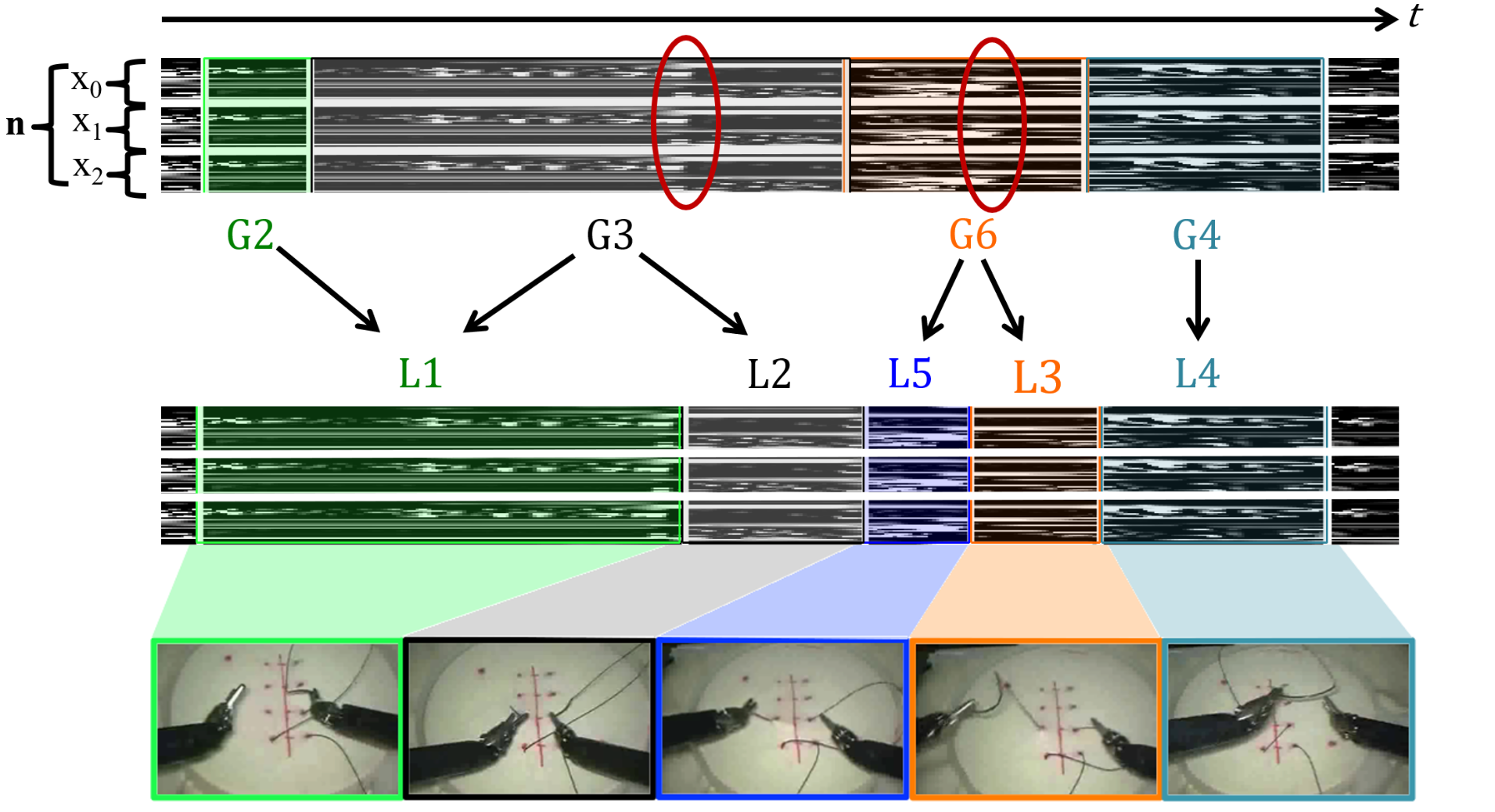}
	\caption[Kinematic data visualization.]{The schematic shows the augmented state vector $\textbf{n}(t)$ = [\textbf{x$_0$} \textbf{x$_1$} \textbf{x$_2$}], \textbf{x$_0$}=\textbf{x}$(t)$, \textbf{x$_1$}=\textbf{x}$(t+1)$, \textbf{x$_2$}=\textbf{x}$(t+2)$, of a surgical demonstration example. Each pixel row corresponds to the value, mapped into gray levels (black = min value, white = max value), of a kinematic feature in time. An overlay of surgeme labels is shown above the raw kinematic values for both the original annotation (G) and the new annotation (L) we propose in this paper. The visual information at each surgeme is also shown below, although we do not use visual features explicitly within our approach.}
	\label{fig:kinem}
\end{figure*}

\subsection{Simultaneous action segmentation and recognition}

We build upon the approach of \cite{Krishnan2018}, treating each demonstration $\textbf{x}(t) \in R^p$, as a realization of a switched linear dynamical system with zero-mean Gaussian process noise $\textbf{w}(t) \in R^p$: 
\begin{equation}
\textbf{x}(t+1)=\textbf{A}_k\textbf{x}(t)+\textbf{w}(t), \quad \textbf{A}_k \in \{\textbf{A}_1,\cdots,\textbf{A}_N\}
\end{equation}
\noindent where each different locally linear regime $\textbf{A}_k \in R^{p \times p}$ corresponds to one of the $N$ different surgemes composing the task. As explained in \cite{Krishnan2018}, under this hypothesis action recognition can be performed by fitting a GMM to the augmented state $\textbf{n}(t)$, defined as:
\begin{equation}
\textbf{n}(t)=[ \textbf{x}(t), \quad \textbf{x}(t+1), \quad \cdots \quad \textbf{x}(t+W)] \in R^{pW}.
\end{equation}
\noindent When $W=1$, GMM fitting is indeed equivalent to solving multiple linear regression problems \cite{Moldovan2015}, one for each action class $\textbf{A}_k$. After model fitting, each trajectory sample is assigned to its most likely mixture component, i.e. its most likely surgeme label.

\subsection{Weakly supervised initialization}

In order to initialize the GMM parameters (mean, covariance and weight of each mixture component), we use a small set of manually-segmented demonstrations. This set is composed of two demonstrations from expert users ($\textbf{e}_1(t), \textbf{e}_2(t)$) and one demonstration from an intermediate user ($\textbf{e}_3(t)$), randomly selected among all demonstrations observed to be free from execution errors (such as needle dropping or multiple attempts of the same gesture). This set provides an exemplary execution of each possible action, which helps to generate a mixture with the correct number of components and appropriate shape.
Exploiting the available ground truth annotations, we fit an initial GMM, denoted as $GMM_0$, to the example demonstrations ($\textbf{e}_1(t), \textbf{e}_2(t), \textbf{e}_3(t)$), thus obtaining the initial values of the mean vector (\boldmath$\mu$\unboldmath$_{0k}$), covariance matrix ($\textbf{C}_{0k}$) and weight ($w_{0k}$) of each mixture component:
$$
\textbf{e}_1(t), \textbf{e}_2(t), \textbf{e}_3(t) \ \xrightarrow{} \ \{\mbox{\boldmath$\mu$\unboldmath$_{0k}$},\ \textbf{C}_{0k},\ w_{0k}\}_{k=1:N}
$$

\subsection{Trajectory segmentation}

Once the initial mixture parameters have been generated, offline segmentation of the full dataset was performed by fitting another GMM onto the unlabelled demonstrations using the Expectation Maximization (EM) algorithm \cite{moon1996}. Each trajectory sample is assigned to its most likely mixture component, i.e. its most likely action label.

\subsection{Ground Truth segmentation redefinition} As introduced in Section B, the action recognition method of this study relies on the hypothesis of local linearity in demonstrations, where each locally linear regime corresponds to a different surgeme.
Thus, we decided to analyse the suturing task in order to check if this hypothesis is approximately verified.

When observing the video records of the suturing demonstrations, we noticed the presence of sudden motion variations during the execution of some of the surgemes. By plotting (Fig. \ref{fig:kinem} top) the corresponding kinematic state vector $\textbf{n}(t)$ as an image 
(where each pixel row corresponds to the value, mapped into gray levels, of a kinematic feature in time), overlaid with ground truth action boundaries (labels G), we noticed indeed the presence of sharp transitions of the kinematic pattern within 
those surgemes (e.g. see G3 and G6). In order to better satisfy the local linearity hypothesis required by our recognition algorithm, we therefore redefined our ground truth annotations (Fig. \ref{fig:kinem} bottom) in a way to avoid abrupt motion variations within surgemes.

\begin{figure}[t]
	\vspace{0.4cm}
	\centering
	{\includegraphics[width=0.8\columnwidth]{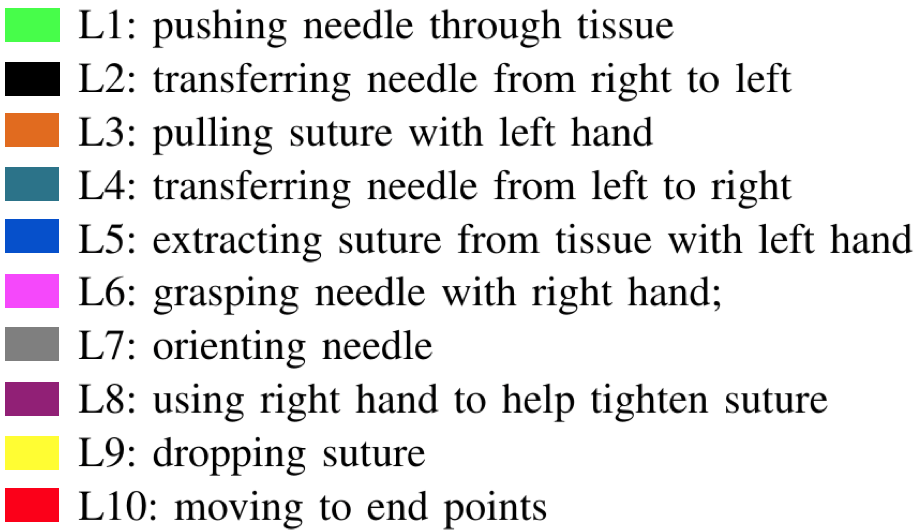}}
	\caption[Legend.]{Redefined action dictionary. Each surgeme is represented with a different colour for visualization purposes.}
	\label{fig:legend}
\end{figure}

Using the video feedback and the original annotations, all the trajectories have therefore been re-segmented according to the following criteria:

\begin{itemize}
	\item Surgeme (G3) \textit{pushing needle through the tissue} is split into (L1) \textit{pushing needle through tissue} and (L2) \textit{transferring needle from right to left}.
	\item Surgeme (G6) \textit{pulling suture with left hand} is split into (L5) \textit{extracting suture from tissue with left hand} and (L3) \textit{pulling suture with left hand}.
	\item Surgeme (G11) \textit{dropping suture and moving to end points} is, when necessary, split into (L7) \textit{orienting needle}, (L9) \textit{dropping suture}, and (L10) \textit{moving to end points}.
	\item Surgeme (G5) \textit{moving to centre of workspace with needle in grip} is most of the times performed simultaneously with (L7) \textit{orienting needle} or (G2) \textit{positioning the tip of the needle}. We therefore include it in either of the two.
	\item Surgeme (G2) \textit{positioning the tip of the needle} and (G3) \textit{pushing needle through the tissue} are merged into (L1) \textit{pushing needle through tissue} class, because there is no clear transition point between the two actions. Moreover, small needle repositioning motions are often performed when inserting the needle through the tissue.
\end{itemize}
\noindent The redefined action dictionary is presented in Fig. \ref{fig:legend}. 

\vspace{0.2cm}
\section{Evaluation metrics}

We evaluate our algorithm performance in a similar way to \cite{Murali2016}. We use both extrinsic metrics, comparing the segmentation result to the ground truth annotations, and intrinsic metrics, measuring the compactness of the generated transition point clusters. 

\subsection{Extrinsic metrics}

\textbf{\textit{Accuracy}}: The accuracy represents the percentage of correctly labelled frames.
\
\vspace{0.2cm}
\textbf{\textit{Normalized Mutual Information (NMI)}}: The NMI measures the alignment between two sequences of labels (X and Y):
$$
NMI(X,Y)=\frac{I(X,Y)}{\sqrt{H(X)H(Y)}} 
$$
\noindent where $I$ is the mutual information and $H$ the entropy \cite{Strehl2002}. This metric is independent of the absolute values of the labels, i.e. the score is not affected by permutations of cluster labels. 

\subsection{Intrinsic metrics}

\textbf{\textit{Silhouette Index (SI)}}: The SI for a single sample is defined as:
$$
s(i)=\frac{b(i)-a(i)}{max\{a(i),b(i)\}} 
$$
\noindent where $a(i)$ is the distance between that sample and the mean of the cluster it belongs to, while $b(i)$ is the distance between that sample and the mean of the nearest cluster it is not part of. We employed the Euclidean distance metric. The Silhouette value is a measure of how similar an object is to its own cluster compared to other clusters. The individual Silhouette value ranges from -1 to +1. We normalised it between 0 and 1, as in \cite{Murali2016}. A high value indicates that the object is well matched to its own cluster and poorly matched to neighbouring clusters. $SI$ is the average Silhouette score over all Ns samples:
$$
SI=\sum_{i=1}^{Ns}{s(i)}
$$
\noindent High $SI$ indicates that the clustering configuration is appropriate.
\noindent We call $SI_{GMM0}$ the SI computed on the clusters identified by our algorithm, and $SI_{GT}$ the SI computed on the Ground Truth clusters.

\vspace{0.2cm}
\section{Experiments and Results}

\begin{figure}[b]
	\centering
	{\includegraphics[width=\columnwidth]{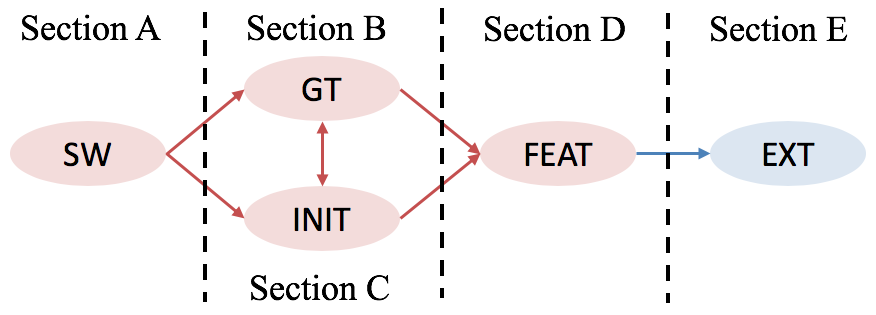}}
	\caption[Experiments.]{We conducted a first set of experiments on expert demonstrations (in red colour), where we first performed a preliminary estimation of the optimal sliding window length (SW, in Section A). We then compared the performance of our initialization method with redefined dictionary to the performance with original Ground Truth annotations (GT, in Section B) and to the performance of K-means initialization method (INIT, in Section C). We finally performed input feature selection (FEAT, in Section D). We then tested, in a second set of experiments, the robustness of our method with redesigned dictionary and selected features on an extended set (EXT, in Section E) of expert, intermediate and novice demonstrations (in blue colour).}
	\label{fig:experiments}	
\end{figure}

As described in Fig. \ref{fig:experiments}, we conducted two sets of experiments, the first on a dataset comprising only expert demonstrations, and the second on a dataset comprising expert, intermediate and novice demonstrations. 

JIGSAWS features 10 suturing demonstrations from expert surgeons, 10 suturing demonstrations from intermediate surgeons and other 20 from novice surgeons. Each demonstration has a different duration of approximately $1105 \pm 432$ frames. The expert demonstrations have generally the shortest duration. 

\begin{table*}[t]
	\vspace{0.4cm}
	\centering
	\footnotesize
	\caption[]{Results on Expert demonstrations.}
	\normalsize
	\label{table:resExp}
	\begin{tabular}{cccccccc}
		\hline
		\multicolumn{2}{c|}{\textbf{Annotations}} & \multicolumn{1}{c|}{\textbf{Original}} & \multicolumn{5}{c}{\textbf{\ \ \ \ \ \ \ \ \ \ \ \ \ \ Proposed}}\\ 
		\hline
		
		\multicolumn{2}{c|}{\textbf{Initialization}} & \multicolumn{1}{c|}{$\textbf{GMM}_0$} & \multicolumn{1}{c|}{\textbf{K-means}} & \multicolumn{4}{c}{$\textbf{GMM}_0$}\\
		\hline
		
		\multicolumn{2}{c|}{\textbf{Signals}} & \multicolumn{3}{c|}{\textbf{All}} & \multicolumn{1}{c|}{\textbf{No pose}} & \multicolumn{1}{c|}{\textbf{No velocity}} & \textbf{No distance}\\
		\hline
		
		& \multicolumn{1}{c|}{\textit{$Accuracy$}} & \multicolumn{1}{c|}{58\%} & \multicolumn{1}{c|}{\textbackslash{}} & \multicolumn{1}{c|}{83\%} & \multicolumn{1}{c|}{55\%} & \multicolumn{1}{c|}{$\textbf{85\%}$} & 77\% \\ 
		\cline{2-8}
		 
		\multirow{-2}{*}{\textit{\begin{tabular}[c]{@{}c@{}}Extrinsic\\ metrics\end{tabular}}} & \multicolumn{1}{c|}{\textit{$NMI$}} & \multicolumn{1}{c|}{0.58} & \multicolumn{1}{c|}{0.58} & \multicolumn{1}{c|}{0.72} & \multicolumn{1}{c|}{0.41} & \multicolumn{1}{c|}{$\textbf{0.73}$} & 0.67 \\ 
		\hline
		
		& \multicolumn{1}{c|}{$SI_{GMM0}$} & \multicolumn{1}{c|}{0.55} & \multicolumn{1}{c|}{0.52} & \multicolumn{1}{c|}{0.57} & \multicolumn{1}{c|}{0.48} & \multicolumn{1}{c|}{$\textbf{0.60}$} & 0.55\\ 
		\cline{2-8} 
		
		\multirow{-2}{*}{\textit{\begin{tabular}[c]{@{}c@{}}Intrinsic\\ metrics\end{tabular}}} & \multicolumn{1}{c|}{$SI_{GT}$} & \multicolumn{1}{c|}{0.59} & \multicolumn{1}{c|}{0.56} & \multicolumn{1}{c|}{0.56} & \multicolumn{1}{c|}{0.54} & \multicolumn{1}{c|}{0.56} & \multicolumn{1}{c}{0.56}\\
		\hline
		
		\\[-1cm]
		\multicolumn{1}{p{1.6cm}}{} & \multicolumn{1}{p{1.6cm}}{} & \multicolumn{1}{p{1.6cm}}{} & \multicolumn{1}{p{1.6cm}}{} & \multicolumn{1}{p{1.6cm}}{} & \multicolumn{1}{p{1.6cm}}{} & \multicolumn{1}{p{1.6cm}}{}
		
	\end{tabular}\\
\end{table*}

\subsection{Sliding window length}
\begin{figure}[t]
	\centering
	{\includegraphics[width=\columnwidth]{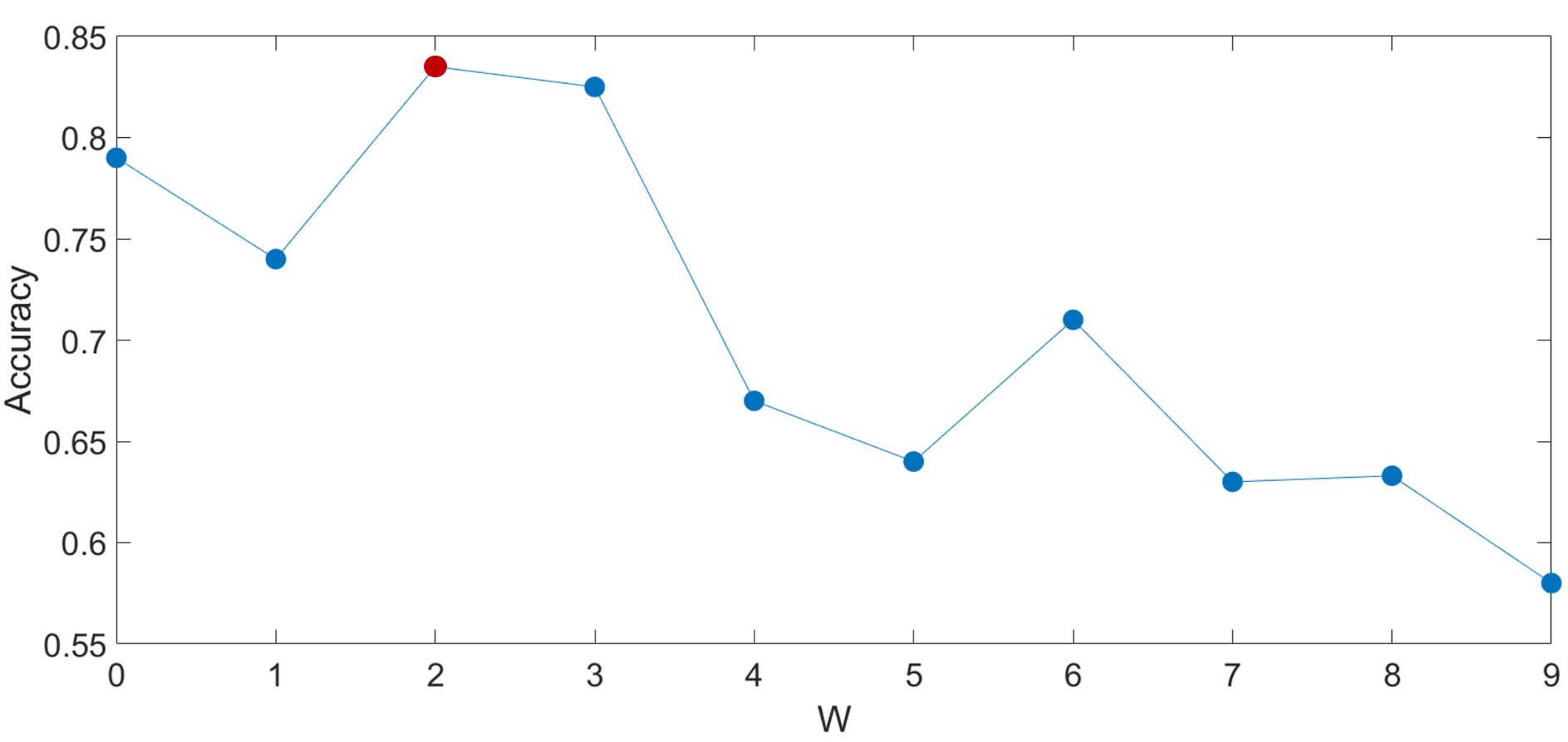}}
	\caption[AccL]{Accuracy score as a function of the sliding window length W. The best recognition performance is obtained with W=2.}
	\label{fig:accl}
\end{figure}

First of all, we conducted a preliminary test to select the optimal sliding window length (W), which defines the dimensions of the augmented feature state $\textbf{n}(t)$. We applied our $GMM_{0}$ method on expert demonstrations for increasing values of W, and computed the corresponding accuracy score with redefined dictionary. As illustrated in Fig. \ref{fig:accl}, the trend is negative, as accuracy decreases when W increases. This could be explained by the corresponding increase in feature dimensionality, which jeopardises the clustering robustness given the same amount of samples. We selected W=2, providing the best recognition performance.

\subsection{Ground truth redefinition}

We then analysed the performance of our $GMM_{0}$ method on both the original and the proposed ground truth annotations. The results are summarized in Table \ref{table:resExp}. With the proposed annotations the extrinsic metrics show remarkable improvement with respect to the original ground truth, with accuracy score increasing of 25\% and NMI score increasing of 14\%.
The $SI_{GMM0}$ is also improved, advancing from 0.55 to 0.57.

\begin{figure}[t]
	\centering
	{\includegraphics[width=\columnwidth]{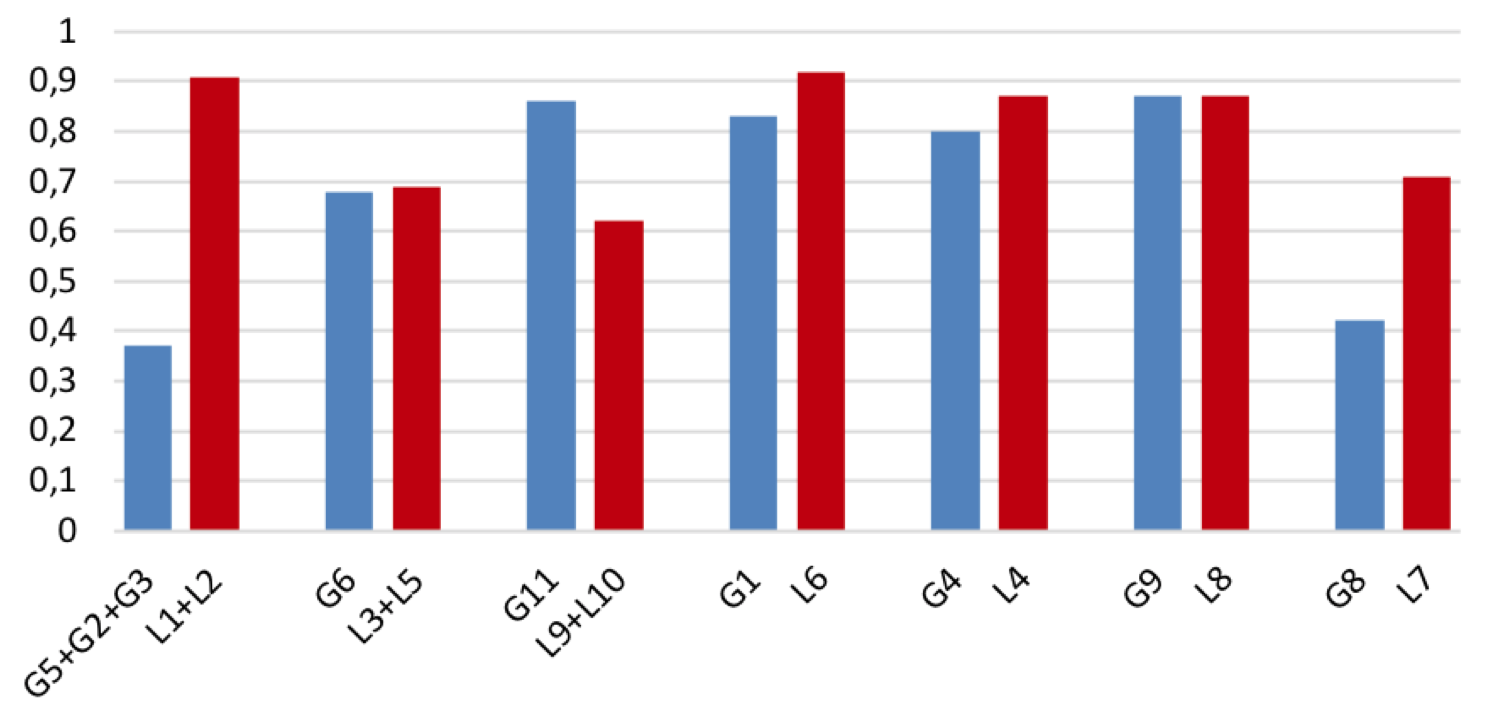}}
	\caption[Correspondent labels - Accuracy.]{Accuracy score comparison between groups of corresponding labels belonging to the original (in blue colour) and the proposed (in red colour) action dictionaries.}
	\label{fig:labelacc}	
\end{figure}

In addition, we compared the recognition accuracy between groups of corresponding labels belonging to the original and the proposed action dictionaries. 
\noindent As shown in Fig. \ref{fig:labelacc}, the fusion of surgeme G2 and G3 into L1 and the separation of surgeme G3 into L1 and L2 give rise to action classes L1 and L2 which can be recognized more robustly, while the separation of G6 into L5 and L3 does not generate significant variations.
Recognition accuracy of G11, split into L9 and L10, decreases with the proposed annotations, but the accuracy of all the other labels (G1, G4, G9, G8) is mostly improved. Overall, a more robust GMM distribution is generated when initialized with the proposed action dictionary.
These results underline the influence of the action dictionary definition on action recognition performance.

\subsection{Initialization technique}

We compared the performance of our $GMM_0$ initialization method with respect to the commonly used K-means initialization method \cite{Arthur2007}, with redefined dictionary (see Table \ref{table:resExp}). The number of K-means clusters is set as the number of action labels in the dictionary and the initial seeds are randomly sampled from the dataset. Being K-means algorithm fully unsupervised, no information about the identity of the generated clusters is available. For this reason the accuracy score was not computed. 

$GMM_0$ initialization leads to 14\% improvement of NMI, as well as increase of $SI_{GMM0}$. K-means indeed assumes equal prior probability for all K clusters (i.e. each cluster has roughly the same number of observations) \cite{Bradley1998} and it is randomly initialized. $GMM_0$, on the other hand, exploits prior information to model the initial location, shape and size of the clusters, leading to more robust action identification.

\subsection{Feature selection}

\begin{figure}[t]
	\vspace{1.5pt}
	{\includegraphics[width=\columnwidth]{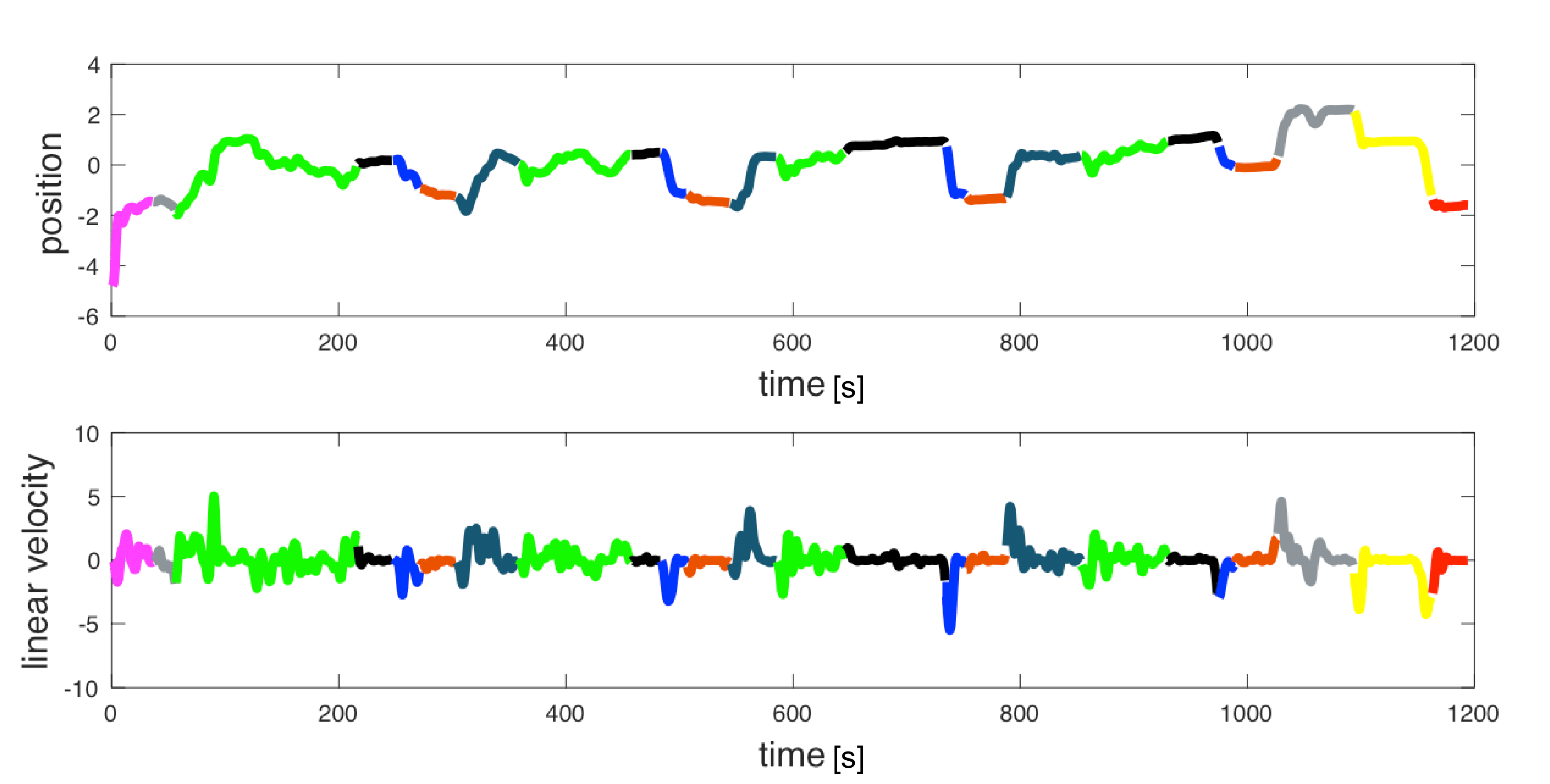}}
	\caption[Position vs velocity signals.]{Example of 
		normalized position trajectory (top) and normalized linear velocity signal (bottom). Each colour represents a different surgeme, as described in Fig.\ref{fig:legend}. Position segments having the same label show higher repeatability than the correspondent velocity segments. Velocity signals are indeed a major source of within-cluster variability.}
	\label{fig:pos_vel}
\end{figure}

We also studied the influence of different signals on the recognition accuracy. Specifically, we analysed the contribution of the pose (NO pose), velocity (NO velocity) and Euclidean distance (NO distance) signals 
by observing how the performance changes in their absence.
The results in Table \ref{table:resExp} suggest that the velocity signals should be discarded, as the recognition performance improves when they are excluded. Velocity signals are indeed a major source of within-cluster variability: not only users with different expertise level perform surgical tasks at different speeds, but even within the same demonstration velocity signals belonging to the same action class show high variability (see Fig. \ref{fig:pos_vel}). The Euclidean distance signals we introduced, instead, give positive contribution to the classification accuracy, as the recognition performance degrades when they are excluded.

\begin{figure}[b]
	{\includegraphics[width=\columnwidth]{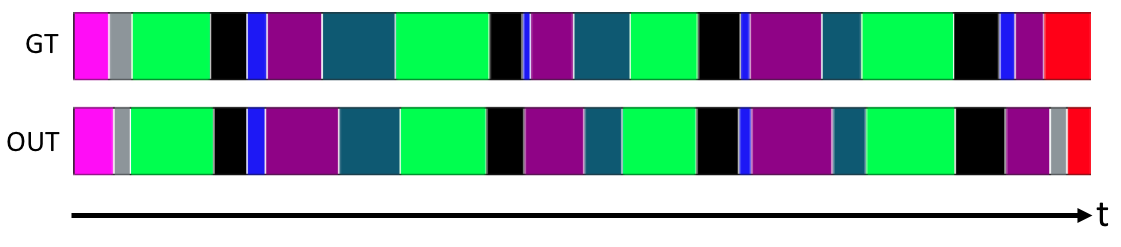}}
	\caption[Example of output.]{Example of segmentation output (bottom) and corresponding ground truth (top). Each colour represents a different surgeme, as described in Fig.\ref{fig:legend}.}
	\label{fig:segm_traj}
\end{figure}

\begin{figure}[b]
	{\includegraphics[width=\columnwidth]{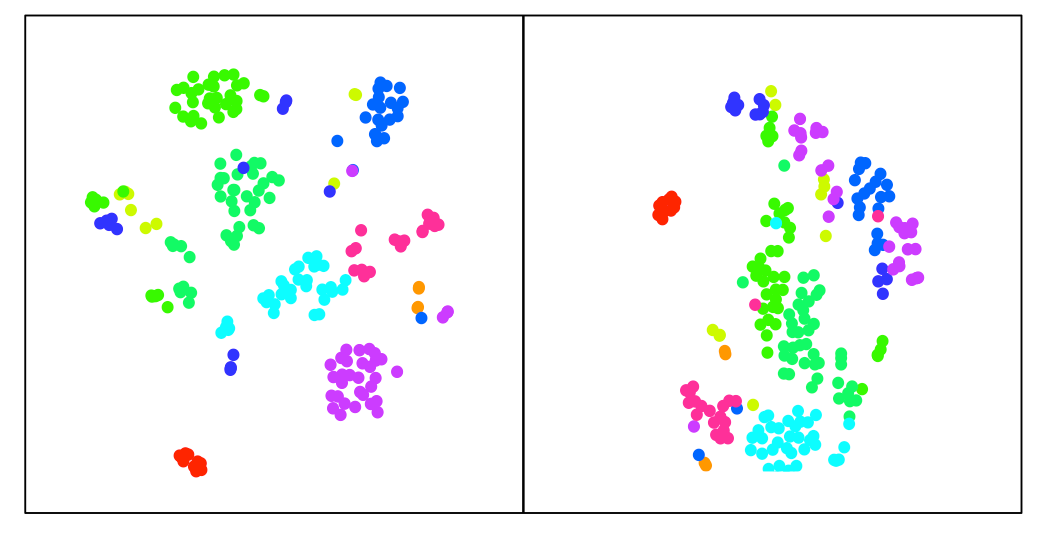}}
	\caption[t-SNE visualization of transition points.]{t-SNE representation of the transition point distribution identified by our algorithm (left), compared to the ground truth distribution (right). Each colour represents a different surgeme, as described in Fig.\ref{fig:legend}.}
	\label{fig:tsne}
\end{figure}

Fig. \ref{fig:segm_traj} shows an example of segmentation output, compared to its ground truth. Fig. \ref{fig:tsne} shows the 2D distribution, obtained with the t-SNE visualization technique \cite{Maaten2008}, of the transition points identified by our algorithm on the expert set. Cluster compactness is visually comparable to the ground truth distribution.


\begin{table}[t]
	\vspace{0.8cm}
	\centering
	\footnotesize
	\caption[Extended dataset.]{Results on extended dataset.}
	\normalsize
	\label{table:extended}
	\begin{tabular}{ccccc}
		\hline
		\multicolumn{2}{c}{Metrics} & \textbf{E} & \textbf{E+I} & \textbf{E+I+N}\\
		\hline
		
		& $Accuracy$ & 85\%	& 77\% & 59\%\\
		\cline{2-5} 
		
		\multirow{-2}{*}{\textit{\begin{tabular}[c]{@{}c@{}}Extrinsic\\ metrics\end{tabular}}}
		& $NMI$	& 0.73 & 0.66 & 0.46\\
		\hline
		
		& $SI_{GMM0}$ & 0.60	& 0.53 & 0.50\\
		\cline{2-5}
		
		\multirow{-2}{*}{\textit{\begin{tabular}[c]{@{}c@{}}Intrinsic\\ metrics\end{tabular}}}
		& $SI_{GT}$ & 0.54 & 0.54 & 0.52\\
		\hline
		
		\\[-1cm]
		\multicolumn{1}{p{0.8cm}}{} & \multicolumn{1}{p{0.8cm}}{} & \multicolumn{1}{p{0.8cm}}{} & \multicolumn{1}{p{0.8cm}}{} & \multicolumn{1}{p{0.8cm}}{}
		
	\end{tabular}
\end{table}

\subsection{Extended dataset}

Finally we extended our algorithm validation to the full dataset, in order to test the robustness of our method, with redesigned dictionary and selected features, against increasing data variability and the presence of spurious motions. Specifically, we used the same initialization ($GMM_0$) as in the previous experiments, but we extended the unsupervised GMM fitting to all expert (E), intermediate (I) and novice (N) demonstrations. As summarized in Table \ref{table:extended}, the lower the expertise level of the surgeon, the lower the final accuracy, NMI and $SI_{GMM0}$ scores. Simple GMM approaches do not exploit temporal constraints such as transition probabilities between actions. This constitutes a major limitation in the analysis of sequential information such as kinematic trajectories, resulting in limited performance as the data variability increases. 


\section{Conclusion}

This paper explored a new weakly supervised approach for surgical gesture recognition, that allows to retain the amount of annotations limited to very few demonstrations. We employed three demonstrations and their ground truth annotations to generate an appropriate initialization for a GMM-based recognition algorithm. Experimental results on real surgical kinematic trajectories during a training exercise confirm that weakly supervised initialization significantly outperforms standard task-agnostic initialization methods. We also demonstrated that recognition accuracy can be improved by carefully designing the optimal channel selection and the appropriate action granularity for the specific task at hand. We believe that inclusion of contextual and semantic information \cite{du2018, allan2018} from video data would further boost the recognition performance \cite{Murali2016}.

However, manual redefinition of the action dictionary based on visual verification still involves a certain degree of subjectivity, and further validation should also be performed to assess the recognition performance for different sets of manually-segmented demonstrations employed in the initialization step. In addition, simple GMM approaches are not robust against increasing data variability. More complex GMM-based methods have been developed specifically for time series analysis, such as Gaussian-HMM \cite{Calinon2010} and GMM-HMM \cite{Tang2010,Loukas2013}. These models introduce transition probabilities between different actions, thus generating a probabilistic action grammar that helps to improve the recognition accuracy. In future work, we will explore the effects of weak supervision on the initialization of transition and observation probability distributions in unsupervised HMM-based approaches. 
Finally, our experiments were conducted on a small-scale dataset, raising concerns about the generalization capability of our approach to surgical data modelling more broadly. More in-depth analysis will be performed on larger and more challenging datasets of robotic surgical demonstrations, e.g. \cite{guru2017}.



\addtolength{\textheight}{-0.3cm}
\bibliographystyle{ieeetr}
\bibliography{library}

\end{document}